%% file: main.tex
\documentclass[
dvipsnames, table,   %
format=sigconf,
anonymous=false,     %
review=false,        %
authorversion=true, %
]{acmart}

\usepackage{custom_commands}
\usepackage{custom_aliases}
\usepackage[normalem]{ulem}
\usepackage{hyperref}

\copyrightyear{2021} 
\acmYear{2021} 
\setcopyright{acmlicensed}\acmConference[GECCO '21 Companion]{2021 Genetic and Evolutionary Computation Conference Companion}{July 10--14, 2021}{Lille, France}
\acmBooktitle{2021 Genetic and Evolutionary Computation Conference Companion (GECCO '21 Companion), July 10--14, 2021, Lille, France}
\acmPrice{15.00}
\acmDOI{10.1145/3449726.3463164}
\acmISBN{978-1-4503-8351-6/21/07}

\begin{document}
\title{How Bayesian Should Bayesian Optimisation Be?}

\author{George {De Ath}}
\email{g.de.ath@exeter.ac.uk}
\orcid{0000-0003-4909-0257}
\affiliation{%
  \department{Department of Computer Science}
  \institution{University of Exeter}
  \city{Exeter}
  \country{United Kingdom}
}

\author{Richard M. Everson}
\email{r.m.everson@exeter.ac.uk}
\orcid{0000-0002-3964-1150}
\affiliation{%
  \department{Department of Computer Science}
  \institution{University of Exeter}
  \city{Exeter}
  \country{United Kingdom}
}

\author{Jonathan E. Fieldsend}
\email{j.e.fieldsend@exeter.ac.uk}
\orcid{0000-0002-0683-2583}
\affiliation{%
  \department{Department of Computer Science}
  \institution{University of Exeter}
  \city{Exeter}
  \country{United Kingdom}
}



\begin{abstract}
  Bayesian optimisation (BO) uses probabilistic surrogate models -- usually
  Gaussian processes (GPs) -- for the optimisation of expensive black-box
  functions. At each BO iteration, the GP hyperparameters are fit to
  previously-evaluated data by maximising the marginal likelihood. However,
  this fails to account for uncertainty in the hyperparameters themselves,
  leading to overconfident model predictions. This uncertainty can be
  accounted for by taking the Bayesian approach of marginalising out the
  model hyperparameters. 
  We investigate whether a fully-Bayesian treatment of the Gaussian process
  hyperparameters in BO (FBBO) leads to improved optimisation performance.
  Since an analytic approach is intractable, we compare FBBO using three
  approximate inference schemes to the maximum likelihood approach, using
  the Expected Improvement (EI) and Upper Confidence Bound (UCB)
  acquisition functions paired with ARD and isotropic \Matern kernels,
  across 15 well-known benchmark problems for 4 observational noise
  settings. FBBO using EI with an ARD kernel leads to the best performance
  in the noise-free setting, with much less difference between combinations
  of BO components when the noise is increased. FBBO leads to
  over-exploration with UCB, but is not detrimental with EI. Therefore, we
  recommend that FBBO using EI with an ARD kernel as the default choice for
  BO.
\end{abstract}

\maketitle

\section{Introduction}
\label{sec:intro}
Bayesian optimisation (BO) is a popular sequential approach for optimising
costly or time-consuming black-box functions that have no derivative
information or closed form \citep{brochu:tutorial:2010, snoek:practical:2012}.
It is a surrogate model-based approach that employs a probabilistic model built
with previous evaluations. BO comprises of two main steps, which are repeated
until budget exhaustion or convergence. Firstly, a probabilistic surrogate
model is constructed, which is typically a Gaussian process (GP) because of
their strength in function approximation and uncertainty quantification
\citep{rasmussen:gpml:2006, shahriari:ego:2016, frazier:tutorial:2018}.
Secondly, an acquisition function is  optimised to select the next
location to expensively evaluate. Acquisition functions combine the surrogate
model's predictions and associated uncertainty to strike a balance
between exploiting areas of design space with good predicted values and
exploring locations with high uncertainty in their predicted values.

During the first step of a BO iteration, the surrogate model must be
learned from the data.
The predominant strategy in BO is to find the model
hyperparameters that maximise the marginal likelihood, also known as the model
evidence. This point-based estimate is known as the \emph{maximum likelihood}
(ML) estimate, or, if we also take into account some prior belief about the 
model's hyperparameters, the \emph{maximum a posteriori} (MAP) estimate.
These are normally found via gradient-based optimisation
\citep{mackay:comparison:1999, gpy:2014, balandat:botorch:2020}. However, the
marginal likelihood landscape may contain multiple optima of similar quality,
as well as flat, ridge-like structures on which  gradient-based optimisers
may get stuck \citep{warnes:likelihood:1987}; a common partial remedy is to take
the best from multiple optimisations from randomly chosen initial
parameters. However, in addition the ML or MAP estimate does not take into
account any uncertainty that exists about the true hyperparameters, leading to naturally overconfident
predictions.

Viewing this from a Bayesian perspective tells us that we need to marginalise
out the hyperparameters of the model; that is, every possible
hyperparameter choice should be weighted
by how well its corresponding model explains the data. Then, we can use the 
prediction of these weighted models to take into account the uncertainty in the
hyperparameters. In all but the simplest of cases, this requires
calculation of  an intractable integral, so in practise approximations to the integral are made using methods such as
variational inference \citep{jordan:vi:1999} and Markov Chain Monte Carlo
(MCMC) \citep{hastings:mh:1970, metropolis:mh:1953, duane:hmc:1987}. Several
works have performed a fully-Bayesian treatment of the hyperparameters in BO,
and some advocate for it to become the prevailing strategy
\citep{osborne:gp:2010, snoek:practical:2012}. Yet most works that apply a
fully-Bayesian approach, \eg 
\citep{benassi:robustei:2011, henrandezlobato:pes:2014, wang:mes:2017} only use
it because it is the \emph{correct} thing to do, without any more
justification. 
Therefore, in this work, we investigate whether a fully-Bayesian treatment
of the surrogate model's hyperparameters leads to improved performance in BO.
Specifically, we investigate the performance of BO using two acquisition
functions (Expected Improvement (EI) and Upper Confidence Bound (UCB)), for
two different Gaussian process kernel types (isotropic and ARD), and under
four different levels of observation noise. These comparisons are made for the
traditional MAP approach and three approximate inference schemes for a
fully-Bayesian treatment.

Our main contributions can be summarised as follows:
\begin{enumerate}
    \item We provide the first empirical study of the effect on BO of a fully-Bayesian 
    treatment of the hyperparameters.
    
    \item We evaluate different combinations of acquisition function, GP kernel
    type, and inference method on fifteen well-known test functions over a
    range of dimensions (2 to 10) and for a range of noise levels.

    \item We show empirically that using the EI with an ARD kernel and
    fully-Bayesian inference using MCMC leads to superior BO performance in the
    noise-free setting.
          
  \item We show that a fully-Bayesian treatment of the hyperparameters
    always leads to (even more) over-exploration with UCB. However, for EI
    a fully Bayesian treatment only increases exploration on
    higher-dimensional functions with increased observational noise level.
\end{enumerate}

We begin in Section~\ref{sec:bo} by reviewing BO and how to perform
fully-Bayesian BO. In Section~\ref{sec:gp} we review GPs, paying
particular attention to the hyperparameter learning, and follow this up in
Section~\ref{sec:approxinf} by reviewing the approximate inference schemes used
in this work. An extensive experimental evaluation is carried out in
Section~\ref{sec:exps}, along with a discussion of the results. We finish with
concluding remarks in Section~\ref{sec:conc}.

\section{Bayesian Optimisation}
\label{sec:bo}
Bayesian optimisation (BO), also known as Efficient Global Optimisation, is a
surrogate-assisted global search strategy that sequentially samples the problem
domain at locations likely to contain the global optimum. It takes into account
both the predictions of a probabilistic surrogate model, typically a Gaussian
process (GP), and its corresponding uncertainty \citep{jones:ego:1998}. It was
first proposed by \citet{kushner:ego:1964}, and improved and popularised by
both \citet{mockus:ei:1978} and \citet{jones:ego:1998}. See
\citep{brochu:tutorial:2010, shahriari:ego:2016, frazier:tutorial:2018} for
comprehensive reviews of BO. Without loss of generality, we can define the
problem of finding a global minimum of an unknown, potentially noise-corrupted
objective function $f : \Real^d \mapsto \Real$ as 
\begin{equation}
    \label{eq:bo:min}
    \min_{\bx \in \mX} f(\bx),
\end{equation}
defined on a compact domain $\mX \subset \Real^d$. In BO it is assumed that $f$
is black-box, \ie it has no (known) closed form and no
derivative information is available. However, we are able to access the results
of its evaluations $f(\bx)$ at any location $\bx \in \mX$. BO is particularly
effective in cases where the evaluation budget is limited due to function
evaluations being expensive in terms of time and/or money. In this case we
wish to optimise $f$ in either as few function evaluations as possible, or as
well as possible for a given budget $T$.

\input{includes/algorithm_bo.tex}

Algorithm \ref{alg:bo} outlines the standard Bayesian optimisation procedure. 
It starts (line~\ref{alg:bo:lhs}) by generating $S$ initial sample locations
$\bX = \{\bx_s\}_{s=1}^S$ with a space-filling algorithm, such as Latin
hypercube sampling \citep{mckay:lhs:1979}. These are expensively evaluated
with the function: $\by = \{y_s \triangleq f(\bx_s)\}_{s=1}^S$. Then, at
each BO iteration, a GP model is usually trained \citep{snoek:practical:2012}
by maximising the log marginal likelihood
$\log p(\by \given \bX, \btheta)$ with respect to the model
hyperparameters $\btheta$, which are usually parameters of the GP kernel,
such as a length scale, and the noise variance assumed to be corrupting the
observed value of $f(\bx)$; see Section~\ref{sec:gp:hypers}.  The marginal 
likelihood may also be multiplied by a prior probability of the parameters,  
$p(\btheta)$, expressing \textit{a priori} beliefs about the parameters.
Maximisation of $\log [p(\by \given \bX, \btheta) p(\btheta)]$ obtains 
the \emph{maximum a posteriori} (MAP)
estimate of the model hyperparameters (line~\ref{alg:bo:train}).
The choice of where to evaluate next in
BO is determined by an acquisition function
$\alpha (\bx \given \by, \bX, \btheta)$, also known as an infill criterion,
which balances the exploitation of good regions of the design space found thus
far with the exploration regions where the predictive uncertainty is high.
Acquisition functions are discussed further in the next section. The 
location $\xnext$ to be expensively evaluated next is selected by 
maximising the acquisition function (line~\ref{alg:bo:xnext}), via
heuristic search, often using an evolutionary algorithm or gradient-based methods, which is possible
because the acquisition function is cheap to evaluate. The selected $\xnext$ is then
expensively evaluated with $f$, the training data is augmented, and the process
is repeated until budget exhaustion.

\subsection{Acquisition Functions}
\label{sec:bo:acq}
Acquisition functions
$\alpha (\bx \given \btheta) : \Real^d \mapsto \Real$ are measures of
quality that enable us to decide which location $\bx \in \mX$ is the most
promising, and thus where we should expend our next expensive evaluation.
They are based on the predictive distribution $p(f \given \bx,  \btheta)$
of the surrogate model, where the dependence on the observed expensive
evaluations ($\bX, \by$) are summarised in $\btheta$.  Acquisition
functions  usually depend on both the posterior mean
prediction $\mu(\bx) = \Expec_{p(f \given \bx, \btheta)}[f(\bx)]$ and its associated uncertainty captured by the
variance $v(\bx) = \mathbb{V}[p(f \given \bx, \btheta) ]$.
Two of the most popular acquisition functions are the Expected Improvement (EI)
\citep{jones:ego:1998} and Upper Confidence Bound (UCB)
\citep{srinivas:ucb:2010}. EI measures expected positive improvement over the
best function value  observed so far $\fstar$:
\begin{equation}
    \alpha_{\text{EI}}(\bx \given \btheta) = 
        \Expec_{p (f \given \bx, \btheta)}
            [\max(\fstar - f(\bx), 0)],
\end{equation}
which can be expressed analytically as \citep{jones:ego:1998}:
\begin{equation}
    \alpha_{\text{EI}}(\bx \given \btheta) = 
        \sqrt{v(\bx)} (s \Phi(s) + \phi(s)),
\end{equation}
where $s = (\fstar - \mu(\bx)) / \sqrt{v(\bx)}$ is the predicted improvement normalised
by its corresponding uncertainty, and $\phi(\cdot)$ and $\Phi(\cdot)$ are the
Gaussian probability density and cumulative density functions respectively.
EI is known to often be too exploitative, resulting in optimisation runs that
can converge prematurely to a local minima \citep{berk:explorationei:2019}.
Various works have tried to curtail this behaviour by increasing the amount of
exploration. \citet{berk:explorationei:2019}, for example, try to do this by
averaging over realisations drawn from surrogate model's posterior distribution
instead of just using the mean prediction. \citet{chen:hei:2021}, like other
authors \citep{sobester:wei:2005, feng:wei:2015}, equate the two terms in EI to
exploitation and exploration and up-weight the amount of the latter
accordingly. However, as shown by \citet{death:egreedy:2021}, only certain
weight combinations allow for this version of EI to be monotonic in both
$\mu(\cdot)$ and $v(\cdot)$; otherwise it can prefer
inferior solutions, \ie with a worse predicted value.

UCB is an optimistic strategy that is the weighted sum of the surrogate model's
posterior mean prediction and its associated uncertainty:
\begin{equation}
    \alpha_{\text{UCB}}(\bx \given \by, \bX, \btheta) = 
        - \left( \mu(\bx) - \sqrt{\beta_t v(\bx)} \right),
\end{equation}
where $\beta_t \geq 0$ is a weight that usually depends on the number of
function evaluations performed thus far and that explicitly controls the
exploration-exploitation trade-off. Setting $\beta_t$ is non-trivial: too small
a value and UCB will get stuck in local optima; too large a value and UCB will
become too exploratory, taking too many evaluations to converge. The convergence
proofs for UCB of \citet{srinivas:ucb:2010} rely on a particular schedule for
$\beta_t$ in which it increases proportional to the logarithm of $t$, although
this scheme has been shown to be over-exploratory for many practical
problems \citep{death:egreedy:2021}.

Other acquisition functions have also been proposed such as \egreedy methods
\citep{death:egreedy:2021}, Probability of Improvement
\citep{kushner:ego:1964}, Knowledge Gradient \citep{scott:kg:2011}, and
various information-theoretic approaches
\citep{hennig:es:2012, henrandezlobato:pes:2014, wang:mes:2017, ru:fitbo:2018}.
However, in this work we focus on EI and UCB due to their popularity.

\subsection{Fully-Bayesian Bayesian Optimisation}
\label{sec:bo:fbbo}
Interestingly, even though fully-Bayesian approaches for Gaussian process (GP)
modelling have been proposed in the literature for several decades, \eg 
\citep{ohagan:curvefitting:1978, handcock:bayesiankriging:1993}, the vast
majority of BO works follow Algorithm \ref{alg:bo}, \ie they perform a MAP
estimate of the GP hyperparameters at each iteration. However, there are some
exceptions to this. \citet{osborne:gp:2010} developed a Bayesian 
approach for global optimisation along with a novel acquisition function, and
showed that their fully-Bayesian approach outperformed the standard MAP
approach. The hugely influential tutorial of \citet{snoek:practical:2012}
advocates a fully-Bayesian treatment of the GP hyperparameters and shows
that it is sometimes superior to MAP estimation. Other works
\citep{benassi:robustei:2011, henrandezlobato:pes:2014, wang:mes:2017} have
also performed a fully-Bayesian approach and have used it to illustrate the
effectiveness of their proposed acquisition functions, rather than specifically
recommending it.

In order to carry out a fully-Bayesian treatment of the hyperparameters in BO,
we need to marginalise out the hyperparameters of the surrogate model, \ie
the acquisition function is averaged weighted by the posterior probability
of the hyperparameters \citep{snoek:practical:2012}:
\begin{equation}
    \label{eq:bo:acq:integral}
    \alpha (\bx \given \by, \bX) = 
        \int \!\! \alpha (\bx \given  \btheta)
                   p(\btheta \given \by, \bX) \,d \btheta,
\end{equation}
where $p(\btheta \given \by, \bX)$ is the surrogate model's 
posterior hyperparameter distribution.
The integral appearing in \eqref{eq:bo:acq:integral} is generally intractable,
but it can be approximated via Monte Carlo integration:
\begin{equation}
    \label{eq:bo:acq:montecarlo}
    \alpha (\bx \given \by, \bX)
        \simeq \frac{1}{M} \sum_{m=1}^M
               \alpha (\bx \given \btheta^{(m)}),
\end{equation}
where $\{\btheta^{(1)}, \dots, \btheta^{(M)}\}$ are samples drawn from
$p(\btheta \given \by, \bX)$ . These samples can be drawn via an
approximate inference method such as Hamiltonian Monte Carlo or
variational inference, both of which are discussed further in
Section~\ref{sec:approxinf}. We note that the MAP estimate of the
hyperparameters can be regarded as approximating $p(\btheta \given \by,
\bX)$ by a delta function that places all the posterior probability
mass at $\btheta_{MAP} = \argmax_{\btheta} p(\btheta \given \by, \bX)$. In using
the estimated integrated acquisition function  \eqref{eq:bo:acq:montecarlo},
the uncertainty in the surrogate model hyperparameters is explicitly taken into
account and may therefore be expected to lead to acquisition of better
locations.

\section{Gaussian Process Surrogates}

\label{sec:gp}
A Gaussian process (GP) defines a prior distribution over functions, such that
any finite number of function values are distributed as a multivariate Gaussian
\citep{rasmussen:gpml:2006}. In GP regression we aim to learn a mapping from a collection of inputs 
$\bX = \{\bx_1, \dots, \bx_n\}$ to their corresponding outputs
$\by = \{y_1, \dots, y_n\}$, where the outputs are often noisy realisations of
the underlying function $f(\bx)$ we wish to model.
Assuming that the observations are corrupted with additive Gaussian noise,
$y = f + \epsilon$ where $\epsilon \sim \Normal(0, \gpnoise^2)$, the
observation model is defined as
$p(y \given f) = \Normal(y \given f, \gpnoise^2)$. In GP regression we place a
multivariate Gaussian prior over the latent variables
$\bff = \{f_1, \dots, f_n\}$:
\begin{equation}
    p(\bff \given \bX, \btheta) \sim \Normal(\bzero, \bK),
\end{equation}
with a covariance $\bK \in \Real^{n \times n}$ with
$K_{ij} = \kappa(\bx_i, \bx_j \given \btheta)$ and associated hyperparameters
$\btheta$. Here, $\kappa(\cdot, \cdot \given \btheta)$ is a positive
semidefinite covariance function modelling the covariance between any pair
of locations. For notational
simplicity, and without loss of generality, we take the mean function of the GP
to be zero; \citep{death:gpmean:2020} discusses BO performance with  different choices of mean
function.

\subsection{Covariance Functions}
\label{sec:gp:cov}
The covariance function $\kappa(\cdot, \cdot \given \btheta)$, also known as a kernel 
(function), encodes prior beliefs about the characteristics of the modelled
function, such as its smoothness. Kernels are typically stationary, meaning
that they are a function of the distance between the two inputs, \ie
$r = \lvert \bx - \xnext \rvert$. One of the most frequently used kernels
is the squared exponential (SE) kernel:
\begin{equation}
    \label{eq:gp:se}
    \kappa_{SE}(\bx, \xnext \given \btheta) = \gpoutputscale^2 \exp 
        \left( - \frac{r^2}{\ell^2} \right),
\end{equation}
with parameters $\ell$ and signal variance $\gpoutputscale^2$ defining its
characteristic length-scale and output-scale respectively.
Using a SE kernel for each input dimension with a separate length-scale
$\ell_i$ results in the SE automatic relevance determination (ARD) kernel:
\begin{equation}
    \label{eq:gp:se-ARD}
    \kappa_{SE}(\bx, \xnext \given \btheta) = \gpoutputscale^2 \exp 
        \left( - \sum_{i=1}^{d} \frac{r_i^2}{\ell_i^2} \right),
\end{equation}
where $r_i = |x_i - x'_i|$.   Allowing
separate length scales for each dimension allows irrelevant dimensions to
be suppressed by placing priors over them which favour large $\ell_i$
\citep{mackay:ard:1992, neal:bayesianlearning:1996}.

It has been argued \citep{stein:interpolation:1999, snoek:practical:2012} that 
the SE kernel has too strong smoothness assumptions for the realistic modelling
of physical processes. Popular alternatives include the \Matern family of
covariance functions \citep{stein:interpolation:1999}.
Here we use the \Matern  kernel with $\nu = 5/2$, as recommended by
\citet{snoek:practical:2012}:
\begin{align}
    \label{eq:gp:matern52}
        \kappa_{Matern}(\bx, \xnext \given \btheta) =
        & \gpoutputscale^2
        \frac{2^{1-\nu}}{\Gamma(\nu)}
        \left(
          \sqrt{2\nu }r
        \right)^\nu
        K_{\nu}\left(
          \sqrt{2\nu } r
        \right),
\end{align}
where $K_{\nu}$ is a modified Bessel function and $ r^2 = \sum_{i=1}^d (x_i
- x_i')^2/\ell_i^2$ is the squared distance between $\bx$ and $\bx'$ scaled by the
length-scales $ \ell_i$ which again allows ARD suppression of irrelevant
dimensions.
Further information on GP kernels
can be found in \citep{rasmussen:gpml:2006, duvenaud:thesis:2014}.

\subsection{Making Prediction with GPs}
\label{sec:gp:prediction}
Given some noisy observations $\by$ at locations $\bX$ and a covariance
function $\kappa(\cdot, \cdot \given \btheta)$, predictions about the 
underlying function $f$ can be made using the GP model. Assuming a Gaussian
noise model, the joint  distribution of the observed training values
$(\bX, \by)$ function values at a test location $(\xpredict, \fpredict)$
is
\begin{equation}
    \label{eq:gp:jointprior}
    \begin{bmatrix*}[l]
        \by \\ \fpredict
    \end{bmatrix*}
    \bigg\rvert \, \bX, \btheta, \gpnoise
    \sim \Normal\left( 
        \bzero, \begin{bmatrix*}[c]
            \bkappa(\bX, \bX \given \btheta) + \gpnoise^2 \bI 
                & \bkappa(\bX, \xpredict \given \btheta)  \\
            \bkappa(\bX, \xpredict \given \btheta)^\top 
                & \kappa(\xpredict, \xpredict \given \btheta)
        \end{bmatrix*}
    \right),
\end{equation}
where the elements of the $n$-dimensional vector $\bkappa(\bX, \xpredict \given \btheta)$ 
are 
$[\bkappa(\bX, \xpredict \given \btheta)]_i = \kappa(\bx_i, \xpredict \given \btheta)$.
Hereafter, the observation noise $\gpnoise$ is incorporated into
$\btheta$ so that $\btheta$ represents both the kernel hyperparameters and the
likelihood noise. We also drop the
explicit dependence of the kernel on $\btheta$  for ease of exposition.

Conditioning the joint distribution \eqref{eq:gp:jointprior} on the
observations $\by$ yields the  predictive distribution of $\fpredict \given
\xpredict, \by, \bX, \btheta $ as a Gaussian distribution:
\begin{equation}
    \label{eq:gp:pred}
    p(\fpredict \given \xpredict, \by, \bX, \btheta)
    \sim
    \Normal (\mu(\xpredict), v(\xpredict)),
\end{equation}
with mean and variance
\begin{align}
    \label{eq:gp:mean}
    \mu(\bx) & = \bkappa(\bX, \bx)^\top 
               (\bkappa(\bX, \bX) + \gpnoise^2 \bI)^{-1} \by \\
    \label{eq:gp:var}
    v(\bx) & = \kappa(\bx, \bx) - \bkappa(\bX, \bx)^\top 
               (\bkappa(\bX, \bX)+ \gpnoise^2 \bI)^{-1} \bkappa(\bX, \bx).
\end{align}

However, before the GP can be used to make predictions the kernel hyperparameters
and the observational noise must be inferred.

\subsection{Learning Hyperparameters}
\label{sec:gp:hypers}
One of the most useful properties of GPs is that we are able to calculate the
marginal likelihood
$p(\by, \bX \given \btheta) \propto p(\by \given \bX, \btheta)$, otherwise known as the model 
evidence \citep{mackay:thesis:1992}, of data $(\bX, \by)$ for a particular
model defined by a set of hyperparameters. The marginal likelihood may be
found by direct integration of the product of the likelihood function of the latent
variables $\bff$ and the GP prior. 
For numerical reasons, the log marginal likelihood is normally used instead:
\begin{align}
\label{eq:gp:lml}
\begin{split}
    \log p(\by \given \bX, \btheta) = 
    & - \frac{1}{2} \by^\top (\bkappa(\bX, \bX) + \gpnoise^2 \bI)^{-1} \by \\
    & - \frac{1}{2} \log \bigl\lvert (\bkappa(\bX, \bX) + \gpnoise^2 \bI)^{-1} \bigr\rvert
    - \frac{n}{2} \log 2 \pi.
\end{split}
\end{align}
The first term in \eqref{eq:gp:lml} is the data-fit term, \ie how well the
model predicts the observed targets $\by$, while the second corresponds to a
complexity penality that depends only on the magnitude of the covariance
function, and the third is a normalisation constant.

The predominant strategy in BO for inferring the hyperparameters is to find
the value of $\btheta$ that maximises the log marginal likelihood
\eqref{eq:gp:lml}. In doing so, we find the \emph{maximum likelihood} (ML)
parameters
$\btheta_{ML} = \argmax_{\btheta} \log p(\by \given \bX, \btheta)$.
Predictions can then be made by using $\btheta_{ML}$ in \eqref{eq:gp:mean}
and \eqref{eq:gp:var}.

However, we often have some prior beliefs about the hyperparameters. For 
example, if our observations were standardised to have zero mean and unit
variance, then it would be extremely unlikely that the likelihood noise would
be larger than one because then the model would predict the majority of the
observations as noise. These beliefs maybe encoded in a prior distribution
$p(\btheta)$, and can be taken into account in the likelihood evaluation by
multiplying the marginal likelihood by the prior distribution:
\begin{equation}
    \label{eq:gp:map}
    p(\btheta \given \by, \bX) \propto p(\by \given \bX, \btheta) p(\btheta).
\end{equation}
In practice, the logarithm of $p(\btheta)$ is added to \eqref{eq:gp:lml}
to arrive at the log posterior distribution, which is then maximised to find
the \emph{maximum a posteriori} (MAP) parameters $\btheta_{MAP}$. Note here
that if we \textit{a priori} believe that all hyperparameter configurations are
equally likely, then $\btheta_{MAP} \equiv \btheta_{ML}$.

\begin{figure}[t] %
\includegraphics[width=\linewidth, clip, trim={7 7 5 5}]
{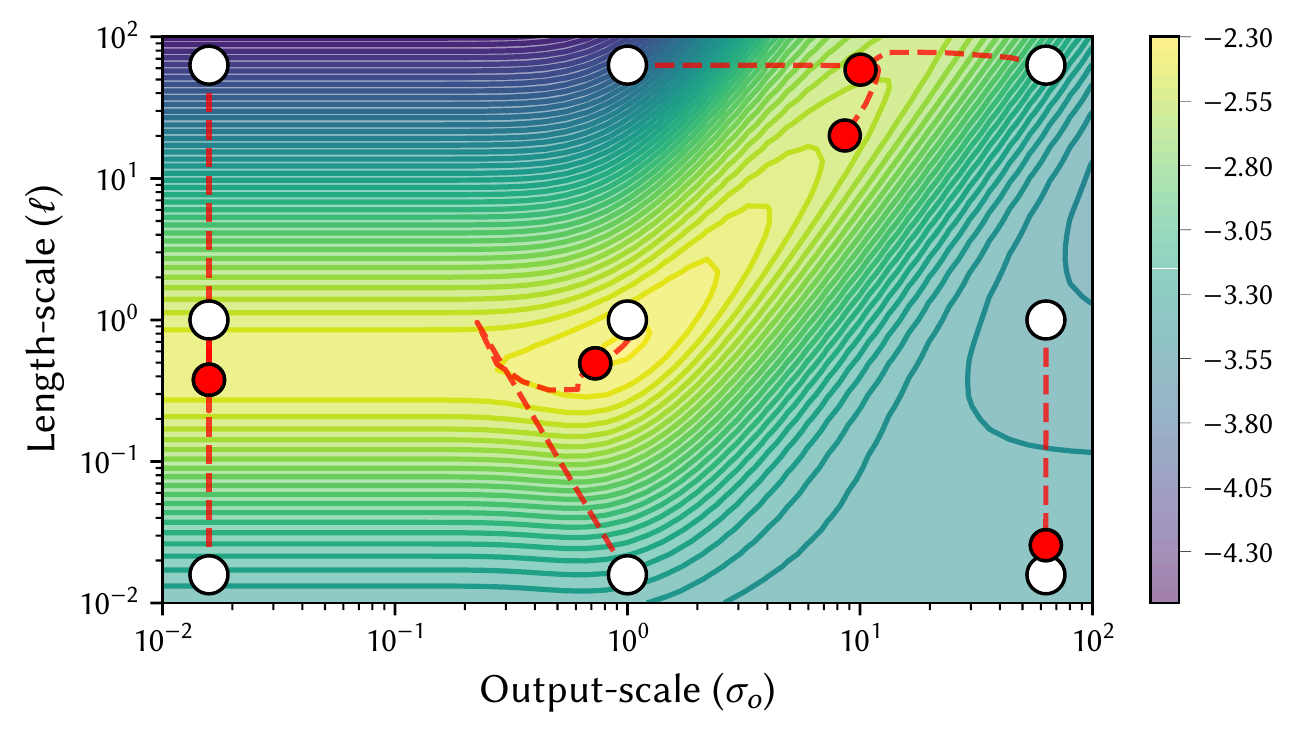}%
\caption{Multi-start gradient-based optimisation of the log marginal likelihood
\eqref{eq:gp:lml}. Starting locations are shown in white, with optimisation
paths shown with red dashed lines and ending locations shown as red circles.
Note how only a few runs successfully find the likelihood peak at $(1,1)$.}
\label{fig:likelihood_optimisation}
\end{figure}
The optimal hyperparameters, $\btheta_{MAP}$ or $\btheta_{ML}$, are
normally found by gradient-based optimisation \citep{mackay:comparison:1999,
gpy:2014, balandat:botorch:2020} with multiple restarts, using \eg L-BFGS-B
\citep{byrd:lbfgs:1995}. These restarts are needed because the landscape may
contain multiple local maxima \citep{rasmussen:gpml:2006}. However, the
gradient-based optimisation can be sensitive to the starting locations because
hyperparameters that are weakly identified can give rise to flat, ridge-like
structures \citep{warnes:likelihood:1987}.

Figure~\ref{fig:likelihood_optimisation} shows nine gradient-based optimisation
runs on the log marginal likelihood \eqref{eq:gp:lml} as a function of the
length-scale $\ell$ and output-scale $\gpoutputscale$. Training data was
generated by drawing a realisation from a GP with an \Matern 5/2 kernel having
hyperparameters $(\ell, \gpoutputscale, \gpnoise)$ = $(1, 1, 0.1)$.
The figure shows the paths taken and final positions (red
circles) by gradient based optimisations starting at the locations
shown as white circles.  Note that $\gpnoise$ was fixed at
$\gpnoise = 0.1$. Only two of the optimisation runs ended up at a location
close to the maximum, with the other runs failing to get close. The three runs
that started on the left-hand side of the figure illustrate the flat, 
ridge-like structures that can occur -- in all three cases the optimiser got
stuck in a suboptimal region of almost zero gradient.

\begin{figure}[t] %
\includegraphics[width=\linewidth, clip, trim={8 7 7 7}]{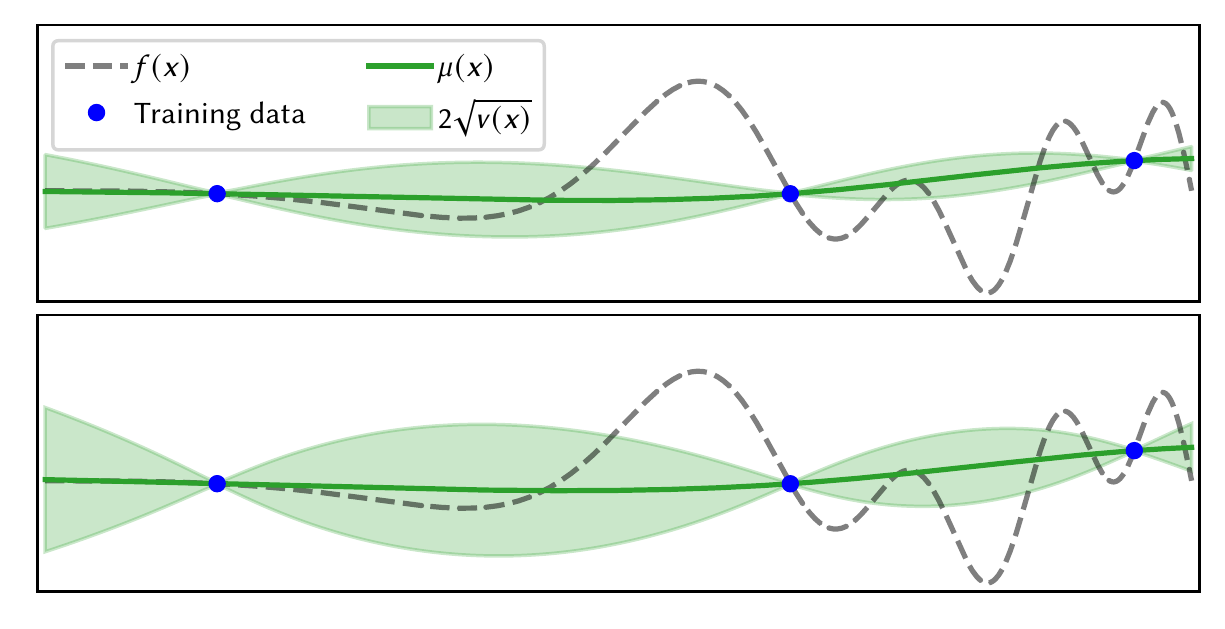}%
\caption{Posterior predictive distributions for two GPs, one fitted with
  the MAP hyperparameter estimate (upper), and one with a fully-Bayesian
  treatment of the hyperparameters via Monte Carlo estimation (lower). Note
  the increased amount of uncertainty in the lower GP as the uncertainty in
  the hyperparameters has been taken into account.}
\label{fig:map_mcmc_comparison}
\end{figure}

As illustrated, point-based estimates of the hyperparameters are subject to
high variability, because they  may become stuck in local minima or on vast plateaux, 
and do not take into account the uncertainty in the hyperparameters
themselves.  Multiple restarts are often sufficient to avoid local minima
and plateaux, but 
accounting for hyperparameter uncertainty requires a fully-Bayesian
formulation.   Given a prior $p(\btheta)$ expressing beliefs about the
hyperparameters, the posterior distribution is given by:
\begin{align}
  \label{eq:bayes-hyperparameters}
  p(\btheta \given \by, \bX) = \frac{p(\by, \bX \given \btheta)
  p(\btheta)}{p(\by, \bX)}.
\end{align}
As noted above, with the posterior distribution on hand, the acquisition
function weighted by the posterior probability of the hyperparameters can be
optimised to find the next location to be expensively evaluated (cf.
\eqref{eq:bo:acq:integral}). However, it is also often of interest to find
the posterior predictive distribution of the function at $\bx$:
\begin{equation}
    \label{eq:gp:intractableposterior}
    p(f \given \bx, \by, \bX) =  \int \!\! p(f \given \bx, \by, \bX, \btheta)
                                        p(\btheta \given \by, \bX) d \btheta.
\end{equation}
The left-hand term of the integrand is a Gaussian
\eqref{eq:gp:pred} and the right-hand term is the hyperparameter posterior.
As illustrated in Figure~\ref{fig:map_mcmc_comparison}, performing a fully-Bayesian
treatment of the hyperparameters in GP regression leads to  increased
uncertainty in the model predictions. This is because we  also account for
the uncertainty in the hyperparameters, rather than  assuming that the MAP
estimate is correct. Similarly, a fully-Bayesian treatment of Bayesian
optimisation accounts for the uncertainty in the hyperparameters when
optimising the acquisition function. 

However, as is frequently the case with Bayesian inference, the integral
necessary to evaluate $p(\by, \bX)$ is intractable, and we therefore turn to
approximate methods.

\section{Approximate Inference}
\label{sec:approxinf}

Practical fully-Bayesian optimisation requires averaging with respect to the GP
hyperparameter distribution. In particular, full account of the model
uncertainty is taken by using the averaged acquisition function
\eqref{eq:bo:acq:integral}. Since this cannot be done analytically, the
averages can be approximated either by simulating draws directly from the
posterior or to approximate the posterior and draw samples from the
approximation; \eg \eqref{eq:bo:acq:montecarlo}.
The former is primarily carried out by Markov Chain Monte Carlo
(MCMC) \citep{metropolis:mh:1953, hastings:mh:1970} methods, and the latter by
many density approximation methods such as Laplace's method, Expectation
Propagation \citep{minka:ep:2001}, and variational inference
\citep{jordan:vi:1999, wainwright:vi:2008}. Compared to MCMC, density
approximation methods tend to be much faster and easier to scale to larger
amounts of data \citep{blei:vireview:2017}, but lack the guarantees of
producing asymptotically exact samples from the target density
\citep{robert:statisticalmethods:2004}. Here, we consider Hamiltonian
Monte Carlo and variational inference as representative methods from the two
main approximate inference styles.

Monte Carlo estimation in the context of GP modelling has been carried
out by a number of authors
\citep{rasmussen:phd:1996, neal:gpmcmc:1997, murray:sliceesampling:2010,
filippone:gpinference:2013, lalchand:gpinference:2020}.  We particularly
note \citet{lalchand:gpinference:2020}, who compared the quality of GP
models using Hamiltonian Monte Carlo and variational inference with ML
estimates.

\subsection{Hamiltonian Monte Carlo}
\label{sec:approxinf:hmc}
Hamiltonian  or Hybrid Monte Carlo \citep{duane:hmc:1987} (HMC), is an MCMC
method which constructs a Markov chain exploring the target probability
density, ensuring that locations are
visited proportional to their probability. HMC is  a promising
method for approximate inference in cases where gradient information is
available \citep{rasmussen:phd:1996}. It is preferred to the more traditional
Metropolis-Hastings (MH) \citep{hastings:mh:1970} algorithm because it is
able to rapidly explore regions of high probability by  introducing auxiliary momentum variables that
are associated with the parameters of the target
density. New proposals as to
where to next move  are generated by simulating Hamiltonian dynamics
for a predefined number of steps $L$, with the dynamics themselves simulated
using a leap-frog symplectic integrator. At
each iteration of the algorithm, the gradients of the log marginal likelihood
of the target density are required for each of the $L$ steps. In the context of
approximate inference for GP regression, this results in the need to invert the
kernel matrix $\bkappa(\bX, \bX \given \btheta)$ $L$ times, hence making HMC a
costly procedure to carry out. See
\citep{neal:hamiltonianmcmc:2011, neal:gpmcmc:1997}
for tutorials on HMC and its application to GPs.

Here, we use a self-tuning variant of HMC known as the No-U-Turn
Sampler (NUTS) \citep{hoffman:nuts:2014}, in which both the path length $L$ and
integration step size are automatically tuned. This avoids the
additional overhead in manually tuning both parameters each time inference is
performed with different data and is thus of particular benefit in BO where
the quantity of data grows at each iteration.

\subsection{Variational Inference}
\label{sec:approxinf:vi}
Variational Inference (VI),  tries to find a most similar approximate
density to a given probability density function \citep{blei:vireview:2017}. 
More formally, we first assume a parametrisable family $\mQ$ of approximate
densities that captures features of the target
density $p(\btheta )$. Then, we try to find $q^*(\btheta) \in \mQ$
that minimises the Kullback-Leibler (KL) divergence to the target
density, $p(\btheta \given \data)$ for data $\data$: 
\begin{equation}
    \label{eq:vi:kl}
    q^*(\btheta) = \argmin_{q(\btheta) \in \mQ}
                   \text{KL} ( q(\btheta) \,||\, p( \btheta \given \data)).
\end{equation}
Finally, we can approximate the posterior with $q^*(\btheta)$ and draw
arbitrarily many samples from it to perform approximate inference.

There any many choices for the family $\mQ$ of approximation, although one of
the key ideas behind VI is to choose $\mQ$ to be expressive enough to model the
target density well, but simple enough to allow for efficient optimisation
\citep{blei:vireview:2017}. A few examples include mean-field (MFVI) and
full-rank (FRVI) Gaussian approximations, as well as the more recent,
expressive, and computationally expensive normalising flows
\citep{rezende:normalizingflows:2015}.

In this work, we focus on MFVI and FRVI because these have been empirically
shown to be suitable for approximating the GP hyperparameter posterior 
distribution \citep{lalchand:gpinference:2020}, while still being relatively
cheap computationally. The mean-field approximation defines $q(\btheta)$ as
a product of independent densities.  Here we choose $q(\btheta)$ to be a
product of normal densities, one for each hyperparameter: 
\begin{equation}
    \label{eq:vi:mfvi}
    q(\btheta ; \bxi_{mf}) 
        = \Normal \big( \btheta \given \bmm, \diag(\bsigma^2) \big)
        = \prod_{j=1}^J \Normal (\theta_j \given m_j, \sigma^2_j ),
\end{equation}
where
$\bxi_{mf} = (m_1, \dots, m_J, \omega_1, \dots, \omega_J) \in \Real^{2J}$
is a vector of unconstrained variational parameters,
$\log(\sigma^2_j) = \omega_j$ and $J = |\btheta|$ is the number of parameters
in the target distribution. Of course, because this is a diagonal approximation
to the true posterior, the mean-field approximation will not capture
correlation between parameters. In contrast to this, the full-rank Gaussian
approximation allows for cross-covariance terms to be modelled explicitly. To
ensure that the learned covariance $\bSigma$ is always positive semidefinite, 
the covariance matrix is written in terms of its  Cholesky factorisation,
$\bSigma = \bL \bL^\top$. This results in an approximating distribution defined
as
\begin{equation}
    \label{eq:vi:frvi}
    q(\btheta; \bxi_{fr}) = \Normal(\btheta \given \bmm, \bL \bL^\top),
\end{equation}
where $\bxi_{fr}= (\bmm, \bL) \in \Real^{J+J(J+1)/2}$ is the vector of
unconstrained variational parameters.

VI seeks to minimise the KL divergence between the target density
$p( \btheta \given \data )$, and the approximating distribution
$q(\btheta)$. It can be shown (\eg \citep{blei:vireview:2017}) that the KL
divergence is minimised by maximising the variational free energy or evidence
lower bound (ELBO),
\begin{equation}
    \label{eq:vi:elbo}
    \text{ELBO}(q) = \Expec_q[\log p( \btheta, \data)] 
                     - \Expec_q[\log q (\btheta)] .
\end{equation}
Both terms in ELBO are readily computed, although the model-specific
computations required to find the gradient with respect to the parameters
$\bxi$ for gradient-based optimisation may be cumbersome. More recently, 
however, the ubiquity of automatic derivative calculations have led to the
development of scalable VI algorithms, namely automatic differentiation VI
(ADVI) \citep{kucukelbir:advi:2017}. ADVI first transforms the inference problem
into one with unconstrained real-values by, for example, taking the logarithm
of parameters that need to be kept strictly positive. It then recasts the
gradient of the ELBO as an expectation of $q$ instead, allowing for the use of
Monte Carlo methods to perform gradient approximation where needed. Lastly, it
reparameterises other gradients in terms of a standard Gaussian, again allowing
for the expectation of a gradient to be calculated (easier) instead of the
gradient of an expectation (hard to impossible); this is known as the
reparameterisation trick \citep{diderik:reparamtrick:2014}.

\section{Experimental Evaluation}
\label{sec:exps}
\label{sec:results:synthetic}
\begin{table}[t]
\centering
\caption{Benchmark functions and dimensionality ($d$).}
\label{tbl:function_details}
\begin{tabular}[t]{lr p{4pt} lr}
\addlinespace[-\aboverulesep]%
    \cmidrule[\heavyrulewidth]{1-2}%
    \cmidrule[\heavyrulewidth]{4-5}%
    \textbf{Name}  & $d$ && \textbf{Name}  & $d$      \\
\cmidrule[\lightrulewidth]{1-2}\cmidrule[\lightrulewidth]{4-5}
    Branin         & 2   && Ackley         & 5, 10    \\
    Eggholder      & 2   && Michalewicz    & 5, 10    \\
    GoldsteinPrice & 2   && StyblinskiTang & 5, 7, 10 \\
    SixHumpCamel   & 2   && Hartmann6      & 6        \\
    Hartmann3      & 3   && Rosenbrock     & 7, 10    \\
    \cmidrule[\heavyrulewidth]{1-2}%
    \cmidrule[\heavyrulewidth]{4-5}%
\addlinespace[-\belowrulesep]
\end{tabular}
\end{table}
We now compare the performance of using a fully-Bayesian treatment of the
hyperparameters in BO (denoted FBBO), with the standard approach of using the
MAP solution (denoted MAP). The EI, and UCB acquisition functions
(Section~\ref{sec:bo:acq}) are compared for both MAP and FBBO using direct
MCMC, MFVI and FRVI across the 15
well-known benchmark functions listed in
Table~\ref{tbl:function_details}\footnote{Function formulae can be found at:
\url{http://www.sfu.ca/~ssurjano/optimization.html}.}. We investigate
several scenarios to compare FBBO with MAP. Specifically, we consider the
noise-free case, where the function of interest is assumed to not be
significantly corrupted
by noise; for this we fix $\gpnoise = 10^{-4}$. We also look at the case where
the function is corrupted by additive Gaussian noise for three different levels
of noise. Note that, because the functions in Table~\ref{tbl:function_details}
are noise-free, we add stochastically generated noise to their
evaluations to simulate a noisy setting. We modify the functions to have
Gaussian additive noise with a standard deviation that is proportional to the
range $|f|$ of possible function values. Concretely, we estimate $|f|$ by
evaluating $10^6$ Latin hypercube samples (LHS) across the function domain,
finding the maximum $f_{max}$ of these samples, and calculating
$|f| = f_{max} - f_{min}$, where $f_{min}$ is the known minimum of
the function. Therefore, for a given  $\sigma_n$, each
function evaluation is a draw from a Gaussian distribution:
$y = \Normal(f(\bx), (\sigma_n |f|)^2) = f(\bx) + \Normal(0, (\sigma_n |f|)^2)$.
We investigate three noise levels, $\sigma_n \in \{0.05, 0.1, 0.2\}$.
We also compare FBBO and MAP using ARD and isotropic kernels, \ie using
one length-scale for each dimension or using one length-scale for all
dimensions of the problem. One might expect, given the reduction in the number
of hyperparameters in the isotropic case, that the hyperparameter posterior
distribution would be less complex and therefore less likely to benefit from a
fully-Bayesian treatment. Overall, this results in 8 sets of experiments across
the test functions: the noise-free setting and three different amounts of noise,
repeated for the ARD and isotropic kernels.

A zero-mean GP with a \Matern $5/2$ kernel 
\eqref{eq:gp:matern52} was used for all experiments. At each BO iteration,
input variables were rescaled to reside in $[0, 1]^d$, and observations were
rescaled to have zero-mean and unit variance prior to GP inference. Relatively uninformative priors
were used, based on BoTorch recommendations \citep{balandat:botorch:2020}, for
the three types of hyperparameters:
$\ell \sim \Ga(3, 6)$, $\gpoutputscale \sim \Ga(2, 0.15)$, and
$\gpnoise \sim \Ga(1.1, 0.05)$,
where $\Ga(a, b)$ is a Gamma distribution with concentration and rate
parameters $a$ and $b$ respectively. Models were initially trained on
$S = 2d$ observations generated by maximin LHS and then optimised for a further
$200 - S$ function evaluations.  The trade-off
$\beta_t$ between exploitation and exploration in UCB was set using Theorem 1 in
\citep{srinivas:ucb:2010}, which implies that $\beta_t$ increases
logarithmically with $t$.
Each optimisation run was repeated $51$ times from a
different set of LHS samples, and the sets of initial locations were used for all
methods to enable statistical comparison. 
An odd number ($51$) of repeats were chosen to allow for the calculation of the
median fitness value without the need for rounding.
For MAP estimation, the GP hyperparameters  were estimated by maximising
$\log [p(\by \given \bX, \btheta) p(\btheta)]$ using a multi-start strategy
with L-BFGS-B \citep{byrd:lbfgs:1995} and $10$ restarts. In all the approximate
fully-Bayesian methods the acquisition functions were averaged over $M = 256$
posterior samples \eqref{eq:bo:acq:montecarlo}. HMC/MCMC sampling inference
(Section~\ref{sec:approxinf:hmc}) was carried out using
PyMC3~\citep{salvatier:pymc3:2016}, with $4$ chains, discarding the first 
$2048$ samples as burn-in and taking every $50$th sample. We note that this is
significantly more than previous works, 
\eg \citep{henrandezlobato:pes:2014, wang:mes:2017} only performed inference every
$10$ BO iterations, using MCMC to draw $M=200$ samples, discarding only the
first $50$ of them, and taking every $3$rd sample. In the non-MCMC BO
iterations, the authors reused the latest set of samples.
When carrying out variational inference
(Section~\ref{sec:approxinf:vi}), optimisation of the ELBO \eqref{eq:vi:elbo}
was undertaken for a maximum of $40000$ steps or until convergence. Finally,
GP models were built using GPyTorch~\citep{gardner:gpytorch:2018} and the
resulting acquisition functions were optimised using
BoTorch~\citep{balandat:botorch:2020}. Full experimental results are available
in the supplementary material, and all code, initial locations and optimisation
runs are available
online\footnote{\url{https://github.com/georgedeath/how-bayesian-should-BO-be}}.

\begin{figure}[t] %
\includegraphics[width=\linewidth, clip, trim={8 7 7 7}]{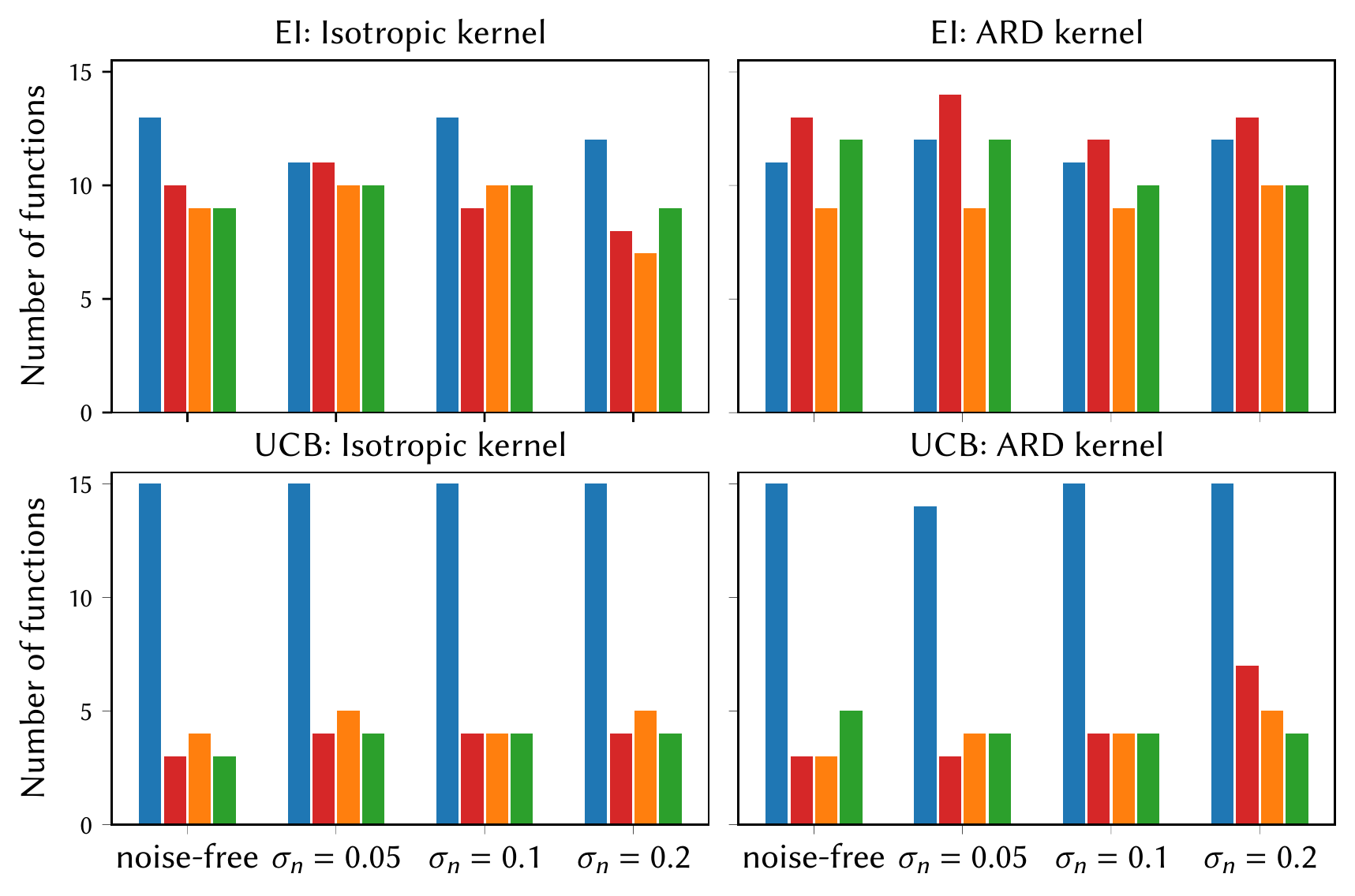}\\
\includegraphics[width=0.65\linewidth, clip, trim={8 29 7 26}]{figs/legend}%
\caption{EI and UCB optimisation summary. Bar heights correspond to the number
of times that an inference method is the best or statistically equivalent to
the best method across the 15 functions, for the 4 noise scenarios and 2 kernel
types.}
\label{fig:acq_comparison}
\end{figure}

Here, we report performance in terms of the logarithm of the simple regret
$R_t$, which is the difference between the true minimum value $f_{min}$ and the
best seen function evaluation up to iteration $t$: 
$\log(R_t) = \log( \min \{f_1, \dots, f_t\} - f_{min})$.

\subsection{Results and Discussion}
\label{sec:exps:results}
Each combination of acquisition function (EI and UCB) and kernel
(isotropic and ARD) were evaluated on the test problems
shown in Table~\ref{tbl:function_details}, for the four different noise levels
($\sigma_n \in \{0, 0.05, 0.1, 0.2\}$).

Figure~\ref{fig:acq_comparison} shows, for the EI and UCB acquisition 
functions and noise levels, the number of times BO with each inference method was the 
best-performing according to a one-sided, paired
Wilcoxon signed-rank test \citep{knowles:testing} with Holm-bonferroni
\citep{holm:test:1979} correction ($p \geq 0.05$). As can be seen from the
plot, for EI with the isotropic kernel, MAP outperforms the other inference
methods. Given that there are only three hyperparameters
$\btheta = (\ell, \gpnoise, \gpoutputscale)$ regardless of the problem
dimensionality $d$, this is not wholly surprising. It is likely that the
hyperparameter posterior distribution \eqref{eq:gp:intractableposterior}
quickly becomes sharply unimodal, particularly given the lack of freedom in the
parameters due the single length-scale. This matches the observations of
\citet{rasmussen:gpml:2006}, who note that as the amount of data increases, as
it does in BO, that one often finds a local optimum that is orders of magnitude
more probable than any other, and, indeed, that it becomes more sharply peaked.

\begin{figure}[t] %
\includegraphics[width=\linewidth, clip, trim={8 7 7 7}]{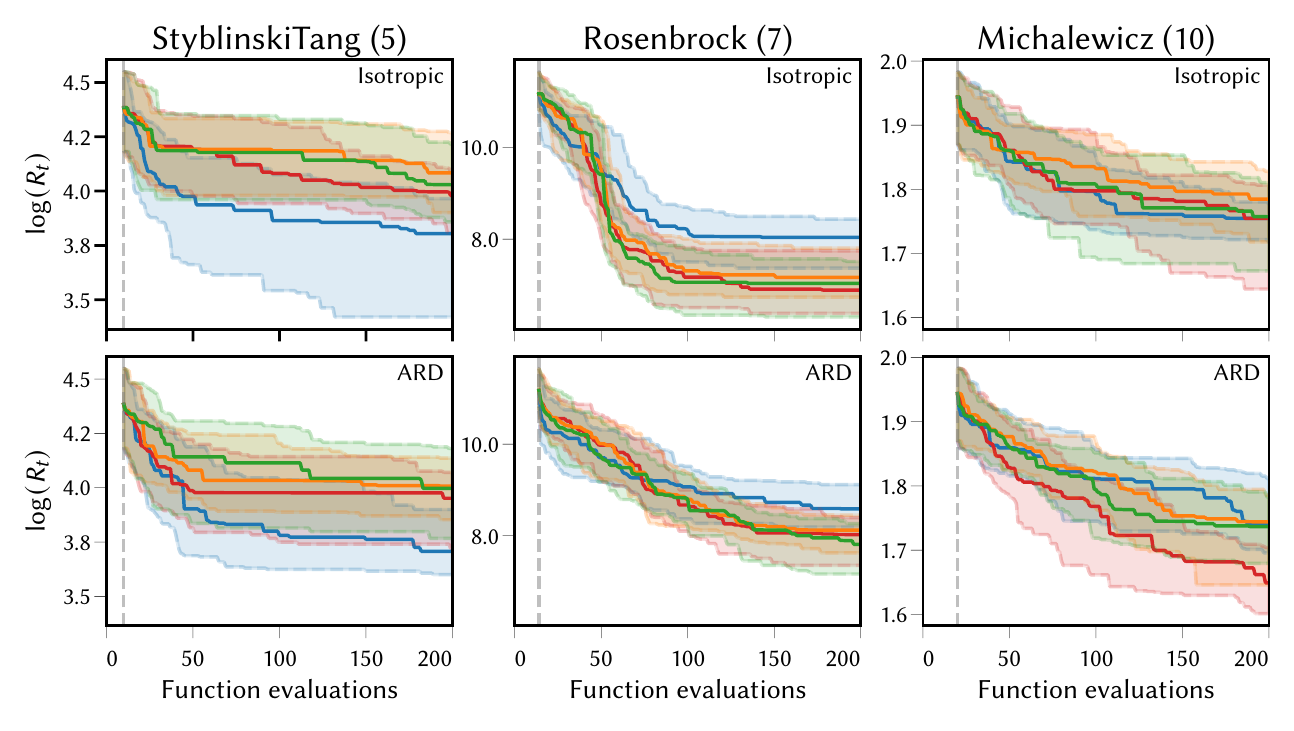}\\
\includegraphics[width=0.65\linewidth, clip, trim={8 29 7 26}]{figs/legend}%
\caption{Illustrative convergence plots for 3 benchmark problems for EI using
an isotropic (upper) and ARD kernel (lower). Each plot shows the median log
simple regret, with shading representing the interquartile range over 51 runs.}
\label{fig:ei_conv_plot}
\end{figure}
Conversely, for EI using an ARD kernel MCMC outperforms MAP. With ARD
there are $2 + d$ hyperparameters, and thus much more freedom to allow
for different, but  equally likely, explanations for the data.
This corresponds to a much more diffuse, and potentially more multimodal,
posterior density. We note that this does not conflict with
\citeauthor{rasmussen:gpml:2006}'s observation because the curse of
dimensionality means that much more data
would be required for 
one mode to become dominant and sharply-peaked.
Figure~\ref{fig:ei_conv_plot} shows some illustrative
convergence plots using EI with an isotropic (upper) and ARD kernel (lower).

Figure~\ref{fig:acq_comparison} also shows that for EI,
BO with MCMC inference (top row, red bars) was often statistically
significantly better than both MFVI and FRVI (orange and green bars respectively)
on the majority of test problems,
kernels, and noise levels. In fact, when MCMC was equal to or better than MAP,
the ordering of the inference methods was always MCMC $\geq$  MAP > FRVI > MFVI. This
suggests that the posterior hyperparameter distribution is not well
approximated by either VI method.  Figure~\ref{fig:likelihood_optimisation}
lends weight to this interpretation because the posterior distribution was
boomerang-shaped, which cannot be well approximated by either the mean
field approximation, in which each parameter is independent, or the full-rank
approximation, which is constrained to be Gaussian.  Note that the
greater flexibility of FRVI allows it to perform better than MFVI.  Due to
the superiority of MCMC over the variational methods,
we compare MAP with MCMC for the remainder of this work.

\begin{figure}[t] %
\includegraphics[width=\linewidth, clip, trim={8 7 7 7}]
{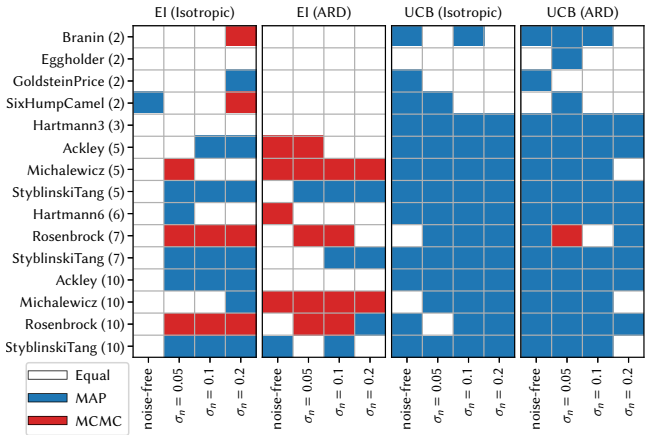}%
\caption{MAP vs. MCMC summary. The colour of each cell indicates whether both
inference methods were statistically indistinguishable from one another
(white), MAP outperformed MCMC (blue) or MCMC outperformed MAP (red).}
\label{fig:map_mcmc_squares}
\end{figure}
Figure~\ref{fig:map_mcmc_squares} summarises the performance of MAP vs. MCMC
for each combination of acquisition function and kernel (columns), and test
problem (rows). As can be seen from the figure, when using UCB, MCMC-based
inference is almost always worse. We suspect that the increased uncertainty
of the full posterior (MCMC) leads to increased exploration with an already
exploratory acquisition function, which ultimately hampers convergence.
Conversely, using EI with an isotropic kernel, neither MAP nor MCMC is
consistently superior. MCMC is almost always better with ARD when using EI,
probably due to the better representation of the posterior hyperparameter
distribution, which allows better prediction and exploration of the function being
optimised. 

In order to investigate whether UCB and EI are really more exploratory when
using MCMC compared with MAP, we calculated the Euclidean distance between
consecutively evaluated locations in each optimisation run, and found that this
is indeed the case -- for plots of the distances see the supplementary
material. With UCB, the distances between consecutive locations are the
smallest for MAP and increase substantially for the other inference methods.
Distances between consecutively evaluated locations for EI, however, did not
follow a similar trend:  In the noise-free setting, there was practically no
difference in the behaviour of EI with MAP or MCMC. Conversely, as the noise
level increases, the approximate inference-based methods become more
exploratory than the MAP estimates, particularly on the higher-dimensional
problems. This indicates that the MAP estimates are sufficient for the
lower-dimensional problems where the hyperparameter posterior distribution
quickly becomes sharply peaked. EI is known to be too exploitative, and, as
discussed in Section~\ref{sec:bo:acq}, several works have tried \emph{ad hoc}
schemes to increase the amount of exploration it performs. Therefore, we argue
that if the EI acquisition function is being used, then taking into account the
hyperparameter uncertainty via a fully-Bayesian treatment of them is essential,
particularly in higher dimensions. Doing so in a principled, Bayesian way leads
to the incorporation of an increased amount of exploration when needed, and
does so without recourse to \emph{ad hoc} rules.

Finally, we compare the different combinations of acquisition function,
inference method, and kernel type to give some general advice to
practitioners as to which of these BO components should be used. As shown
in the supplementary material, EI with a fully-Bayesian treatment of the
hyperparameters and an ARD kernel leads to the best performance in the
noise-free setting, with EI vastly outperforming UCB for all combinations of
components. However, as the noise level increases, the picture becomes less
clear. UCB improves its performance in comparison to EI and the methods are
roughly equal for the different noise levels (supplementary material,
Figure~3). EI's deteriorating performance at higher noise levels may indicate
that the surrogate model is poorer at representing the noisy function, leading to
insufficient exploration. This is in contrast to the more exploratory UCB,
which over-explores in the noise-free case \citep{berk:explorationei:2019}. It
would be interesting to investigate BO strategies for optimisation in the
presence of noise. It is clear that a fully-Bayesian treatment of the
hyperparameters should be avoided with the UCB acquisition function because the
increased level of exploration led to poor performance for both kernel types
across all levels of noise.

\section{Conclusion}
\label{sec:conc}
We  investigated how a fully-Bayesian treatment of the
Gaussian process hyperparameters  affects optimisation performance.
With noise-free evaluations, EI
with a MCMC inference and an ARD kernel was the best combination of evaluated
BO components. However, as the observational noise level increased, there was less to
differentiate between the components. In the case of EI, MCMC was found to be
more effective with an ARD kernel than an isotropic one. We attribute this
to additional flexibility to model the complex and possibly multi-modal
hyperparameter posterior that is afforded by a kernel that treats different
dimensions with different length scales. 
MCMC is generally superior to variational methods (MFVI and FRVI) because the
marginal posterior is often sufficiently complex that it is poorly modelled
with the (necessarily unimodal) mean field or full-rank Gaussian approximations.

We do not recommend the fully-Bayesian approach with UCB because the
additional hyperparameter uncertainty leads to even greater exploration
with the already over-exploratory UCB acquisition function. However, this
is not the case for EI and we, therefore, recommend the fully-Bayesian
treatment of the hyperparameters in BO using MCMC because it allows for a
principled way to increase exploration without any \emph{ad hoc} enhancements to EI.

Other important future directions also focus on the modelling ability of the
surrogate. Particular aspects are the non-stationarity induced as the
optimiser converges \citep[e.g.][]{snoek:inputwarping:2014} and improving the
modelling of functions with degenerate features, such as discontinuities, using
deep GPs \citep{damianou:deepgp:2013, salimbeni:dsdeepgp:2019}.

\begin{acks} 
This work was supported by \grantsponsor{}{Innovate UK}{https://gtr.ukri.org}
[grant numbers \grantnum{}{104400} and \grantnum{}{105874}].
The authors would like to acknowledge the use of the University of Exeter 
High-Performance Computing (HPC) facility.
\end{acks}

\balance %
\bibliographystyle{ACM-Reference-Format}
\bibliography{ref}

\end{document}


\title{Fully-Bayesian Bayesian Optimisation}
\subtitle{Supplementary Material}

\author{George {De Ath}}
\email{g.de.ath@exeter.ac.uk}
\orcid{0000-0003-4909-0257}
\affiliation{%
  \department{Department of Computer Science}
  \institution{University of Exeter}
  \city{Exeter}
  \country{United Kingdom}
}

\author{Richard M. Everson}
\email{r.m.everson@exeter.ac.uk}
\orcid{0000-0002-3964-1150}
\affiliation{%
  \department{Department of Computer Science}
  \institution{University of Exeter}
  \city{Exeter}
  \country{United Kingdom}
}

\author{Jonathan E. Fieldsend}
\email{j.e.fieldsend@exeter.ac.uk}
\orcid{0000-0002-0683-2583}
\affiliation{%
  \department{Department of Computer Science}
  \institution{University of Exeter}
  \city{Exeter}
  \country{United Kingdom}
}

\renewcommand{\shortauthors}{De Ath, Everson, and Fieldsend}

\readlist*\problemsets{noise-free, noise=0.05, noise=0.1, noise=0.2}
\readlist*\problemprintnames{noise-free, $\sigma_n=0.05$, $\sigma_n=0.1$, $\sigma_n=0.2$}
\readlist*\kernels{Isotropic, ARD}
\readlist*\kernelprintnames{isotropic, ARD}
\readlist*\acqfuncs{ei, ucb}
\readlist*\acqprintnames{EI, UCB}

\maketitle
\appendix

\section{Introduction}
In this supplement we include extra results that could not be fit into the main
paper due to space constraints. In the following sections, when comparing
methods, the best method(s) are determined by whether a method either has the
lowest median regret or is statistically indistinguishable from the method
with the lowest median regret according to a one-sided, paired Wilcoxon
signed-rank test \citep{knowles:testing} with Holm-bonferroni
\citep{holm:test:1979} correction ($p \geq 0.05$).

\section{Inference summaries: MAP vs. MCMC}
\label{sec:inf_summary}
\begin{figure}[H]
\centering
\includegraphics[width=0.5\textwidth]{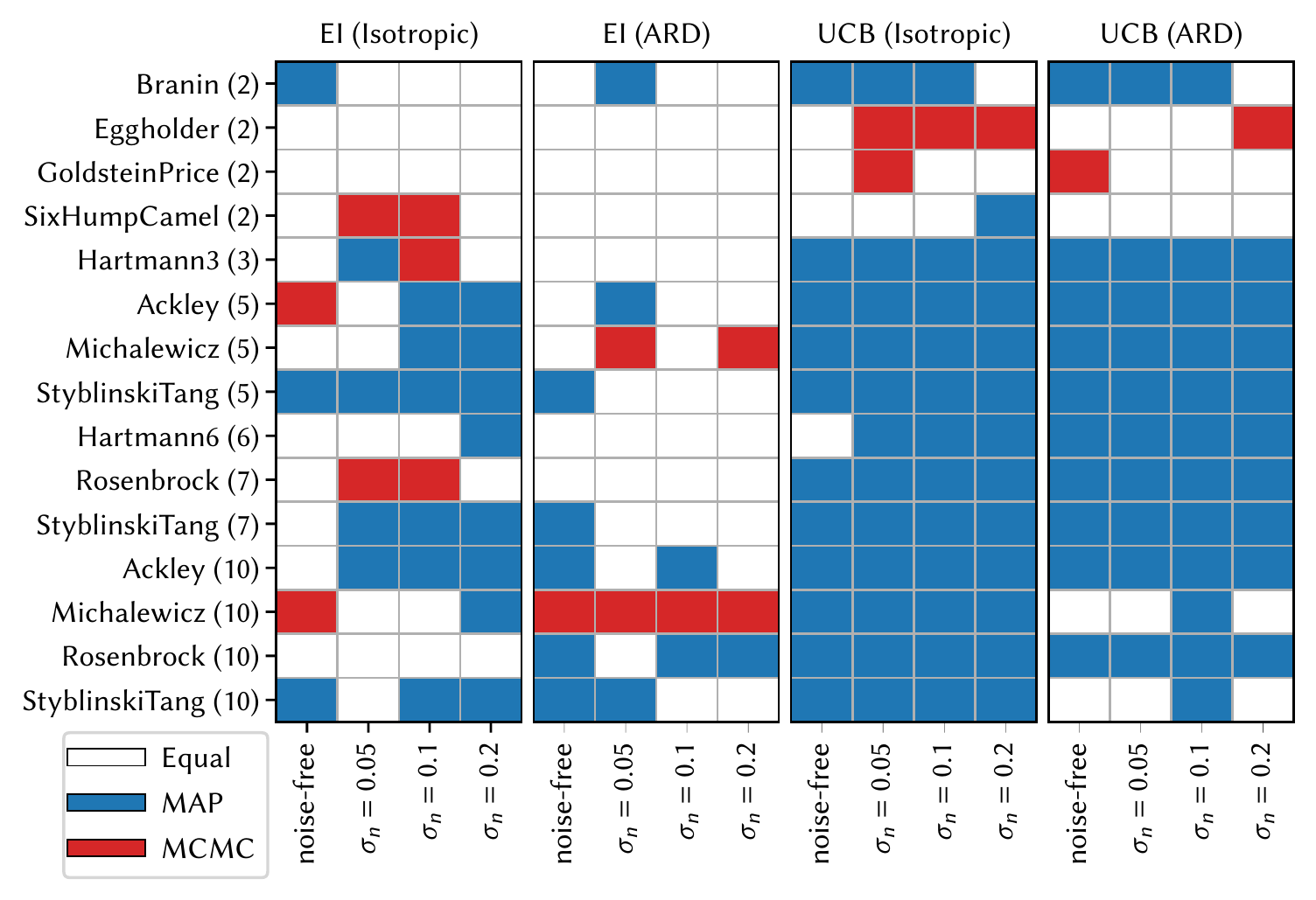}%
\includegraphics[width=0.5\textwidth]{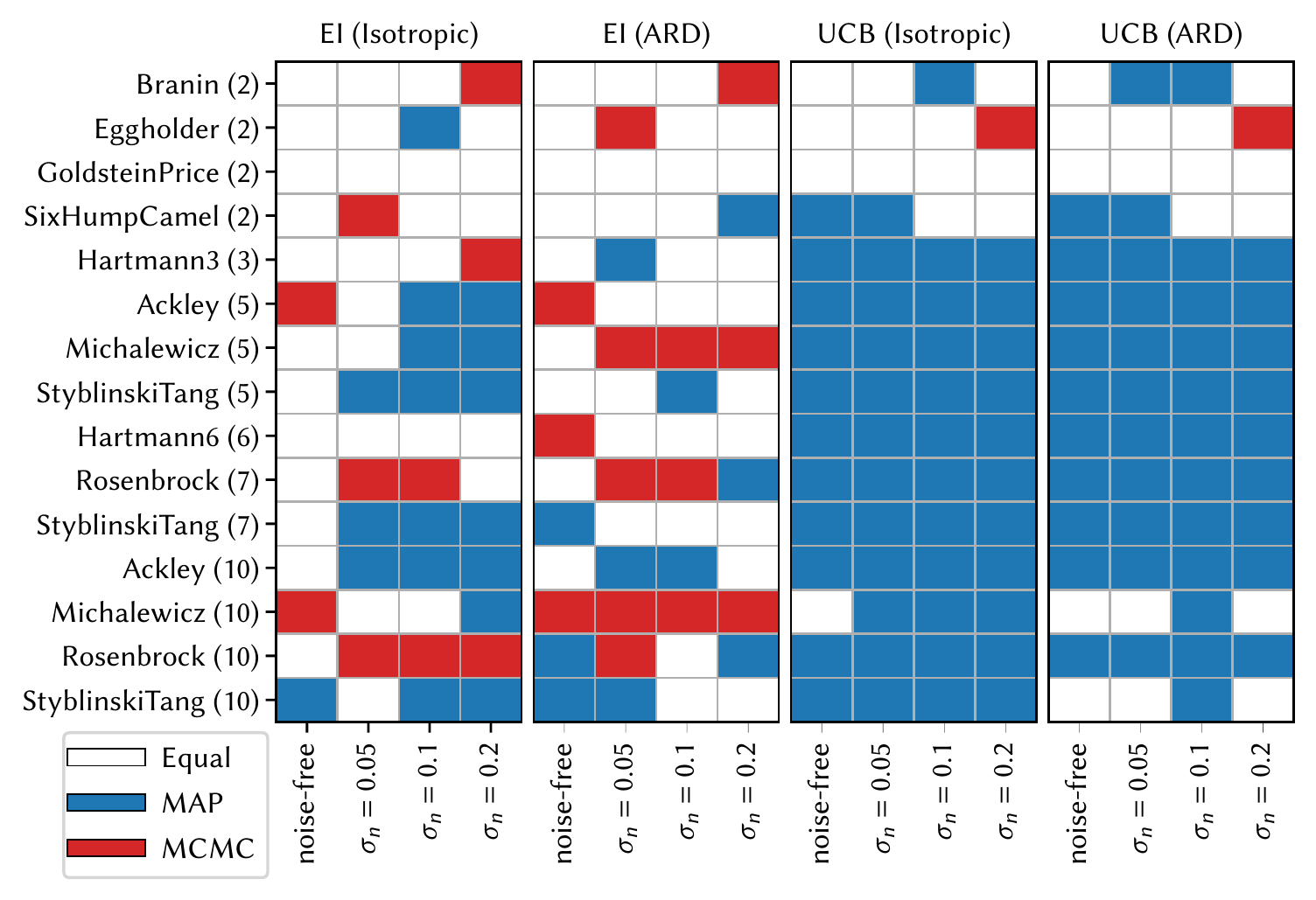}%
\caption{MAP vs. MCMC inference summary after $T = 50$ (left) and $T = 100$
(right) function evaluations. The colour of each cell corresponds to whether
both inference methods were statistically indistinguishable from one another
(white), MAP performed better than MCMC (blue) and MCMC performed better than
MAP (red).}
\label{fig:inference_50-100}
\end{figure}

\begin{figure}[H]
\centering
\includegraphics[width=0.5\textwidth]{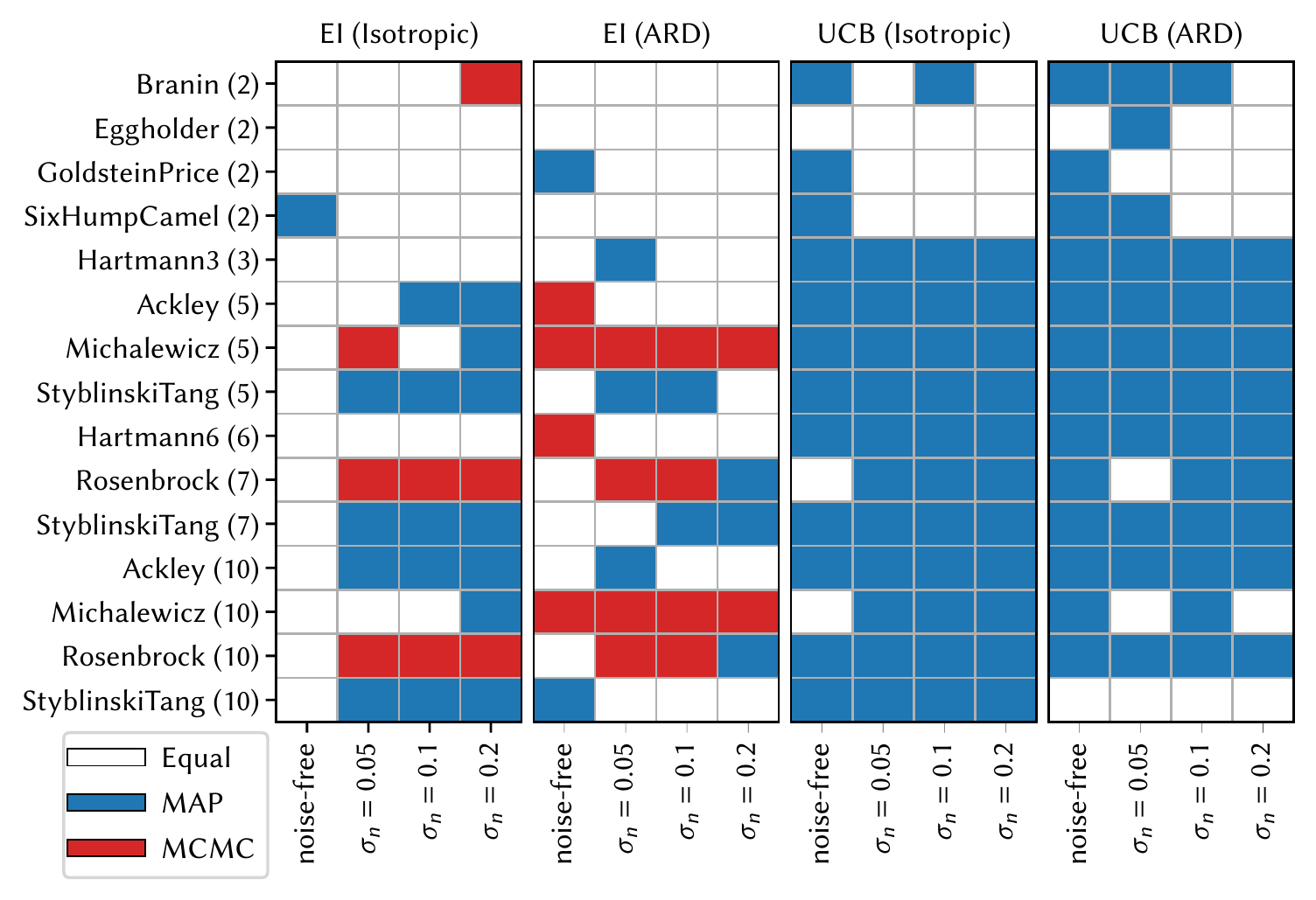}%
\includegraphics[width=0.5\textwidth]{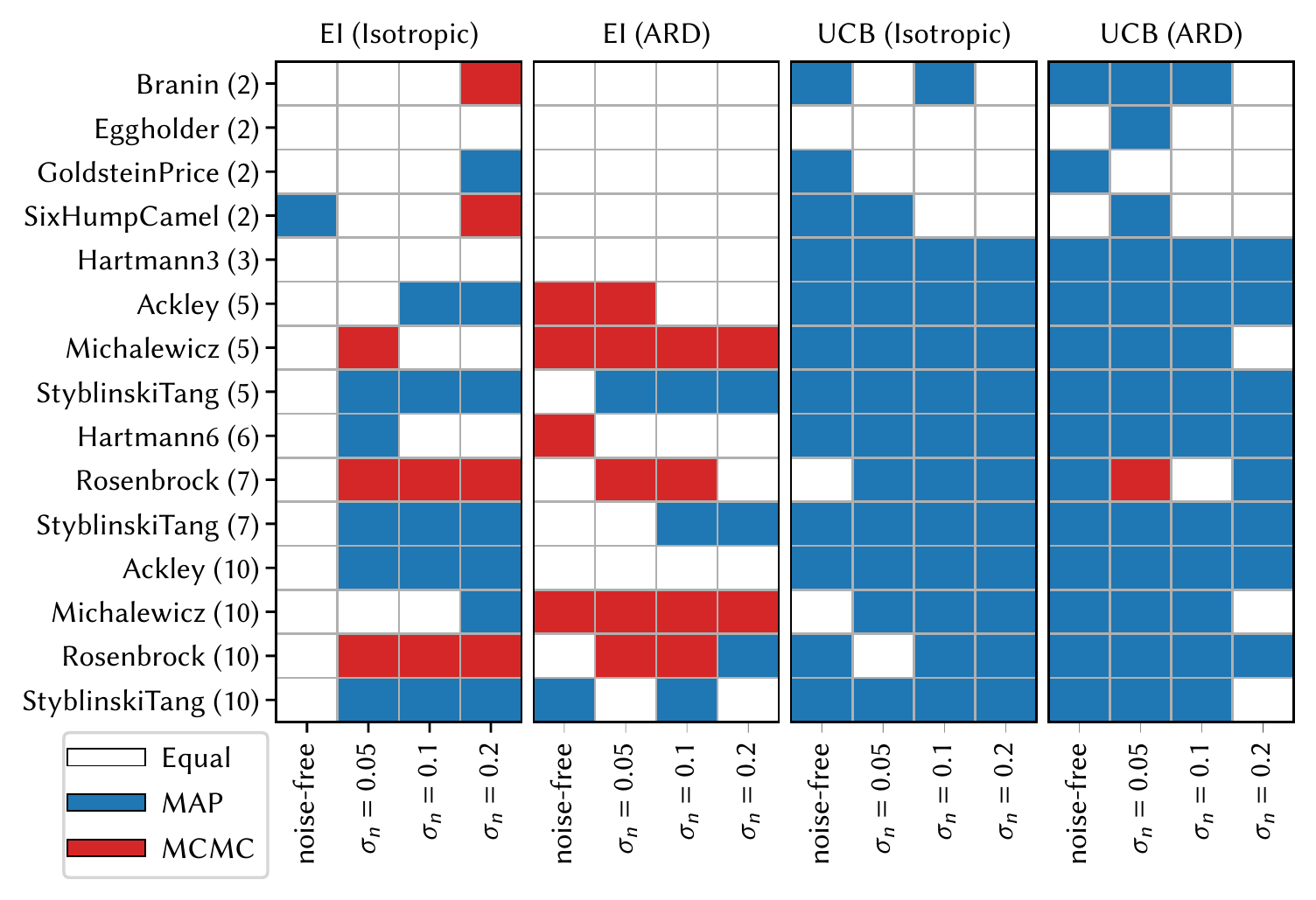}%
\caption{MAP vs. MCMC inference summary after $T = 150$ (left) and $T = 200$
(right) function evaluations. The colour of each cell corresponds to whether
both inference methods were statistically indistinguishable from one another
(white), MAP performed better than MCMC (blue) and MCMC performed better than
MAP (red).}
\label{fig:inference_150-200}
\end{figure}

Here we show the inference summary plots with budgets
$T \in \{50, 100, 150, 200\}$ function evaluations.
Figures~\ref{fig:inference_50-100} and \ref{fig:inference_150-200} summarise
the performance of MAP vs MCMC for each combination of acquisition function and
kernel (columns), and test problem (rows).

\section{Optimisation Summary for Differing Levels of Noise}
\begin{figure}[H]
\centering
\includegraphics[width=\textwidth]{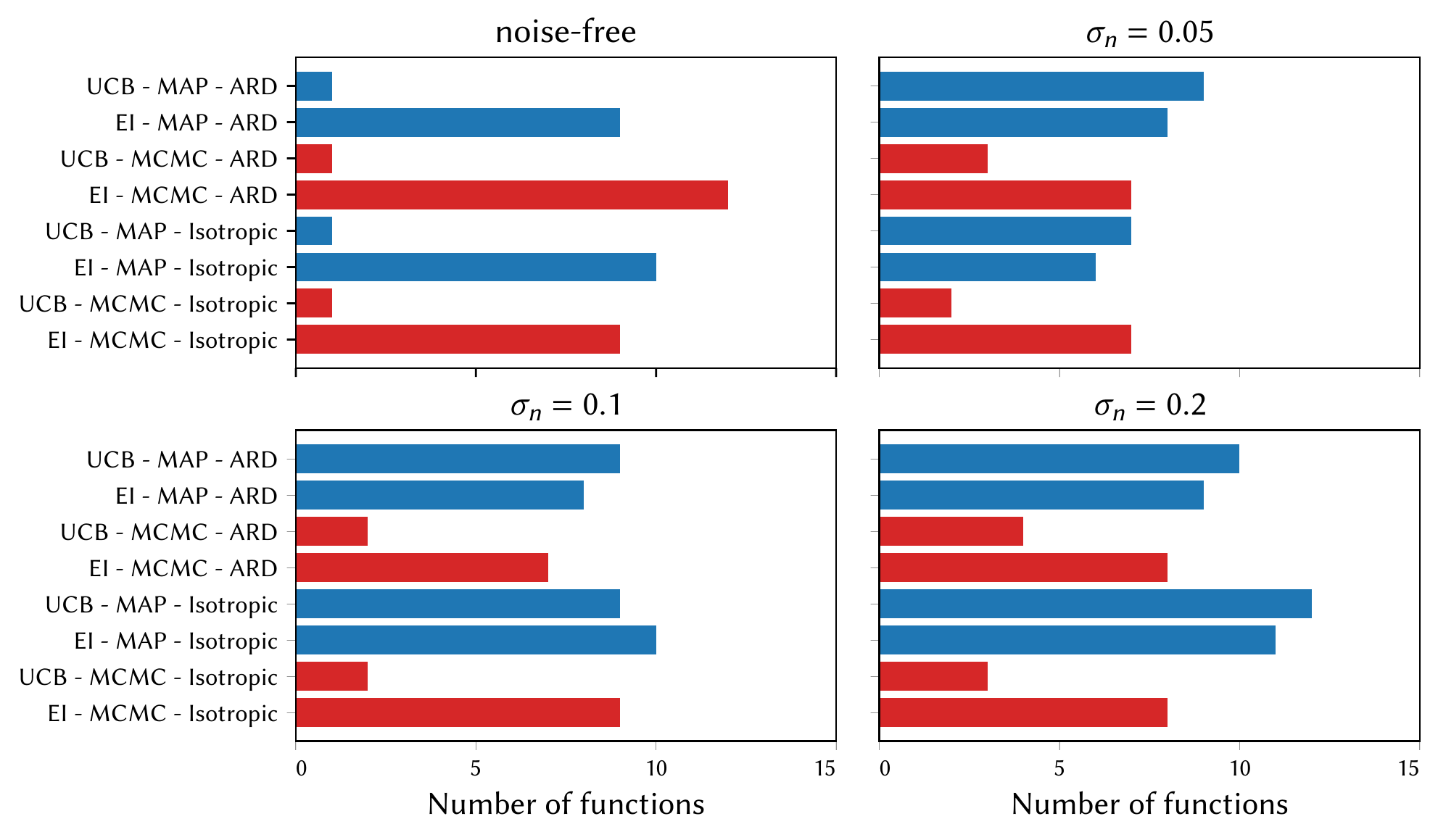}
\caption{Optimisation summary for each level of noise. Bar lengths correspond
to the number of times each combination of acquisition function, inference
method and kernel type was either the best performing or statistically equal to
the best performing combination.}
\label{fig:noise_ranking_summary}
\end{figure}

Figure~\ref{fig:noise_ranking_summary} summarises the performance of each
combination of acquisition function, inference method and kernel type for
each of the four noise settings. As can be seen from the plots, as the noise
level increases, EI becomes less dominant.

\section{Results Tables}
In this section we show the results tables for each of the experiments. The
tables show the median log simple regret as well as the median absolute
deviation (MAD) from the median, a robust measure of dispersion. The method
with the best (lowest) median regret is shown in dark grey, and those that are
statistically indistinguishable from the best method are shown in light grey.
\foreachitem \acq \in \acqfuncs {%
    \foreachitem \kernel \in \kernels {%
        \foreachitem \ps \in \problemsets {%
            \input{includes/\ps_\kernel_\acq.tex}
        }
    }
}

\section{Convergence and Distance plots}
The figures in this section show the convergence and distance plots for each
combination of acquisition function, kernel type and level of function noise.
The convergence plots show the median log simple regret, with shading
representing the interquartile range over 51 runs, and the dashed vertical line
indicating the end of the initial LHS phase. The distance plots show the
normalised Euclidean distance between consecutively selected locations over the
optimisation run. For each $d$-dimensional problem, distances are normalised by
the largest possible distance possible, \ie $\sqrt{d}$, so that distances
reside in $[0, 1]$.
\foreach \i in {1,...,\acqfuncslen} {%
    \def \acq {\acqfuncs[\i]}%
    \def \acqname {\acqprintnames[\i]\xspace}%
    \foreach \j in {1,...,\kernelslen} {%
        \def \kernel {\kernels[\j]}%
        \def \kernelname {\kernelprintnames[\j]\xspace}%
        \foreach \k in {1, ...,\problemsetslen}{%
            \def \ps {\problemsets[\k]}%
            \def \psname {\problemprintnames[\k]\xspace}%
            \begin{figure}[H]
            \centering
            \includegraphics[width=\textwidth]{figs/\ps_\kernel_\acq.pdf}\\
            \includegraphics[width=\textwidth]{figs/distances_\ps_\kernel_\acq.pdf}\\
            \includegraphics[width=0.5\textwidth]{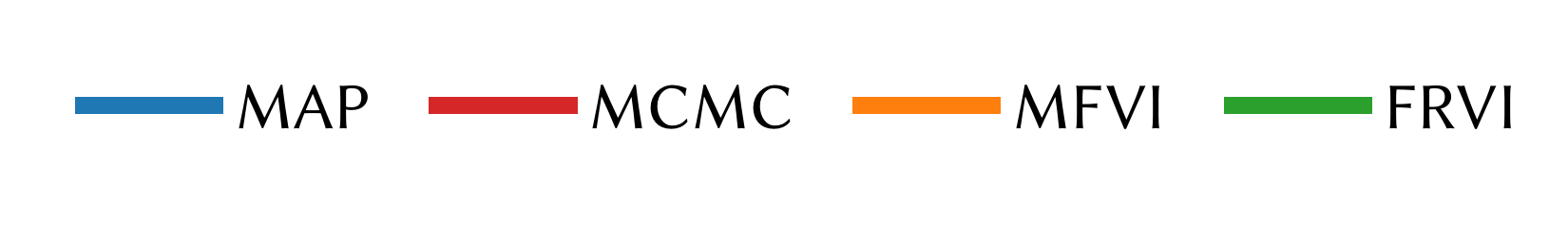}%
            \caption{Convergence (upper) and distance (lower) plots for the
            \acqname acquisition function with an \kernelname kernel on the
            \psname problems.}
            \end{figure}
        }
    }
}

\bibliographystyle{ACM-Reference-Format}
\bibliography{ref}

%% file: includes/algorithm_bo.tex
\begin{algorithm} [t!]
    \caption{Standard Bayesian optimisation.}
    \label{alg:bo}
    \begin{algorithmic}[]
        \State \textbf{Inputs:}
        \State {%
            \setlength{\tabcolsep}{2pt}%
            \begin{tabular}{c p{2pt} l}
                $S$ &:& Number of initial samples \\
                $T$ &:& Budget on the number of expensive evaluations
            \end{tabular}
            }%
    \end{algorithmic}
    \medskip

    \begin{algorithmic}[1]
        \Statex \textbf{Steps:}
        \State $\bX \gets \LatinHypercubeSampling(\mX, S)$
            \label{alg:bo:lhs}
            \Comment{\small{Initial samples}}

        \State $\by \gets \{y_s\triangleq f(\bx_s)\}_{s=1}^S$
            \Comment{\small{Expensively evaluate all initial samples}}

        \For{$t = S + 1 \rightarrow T$}
            \State $\btheta \gets \argmax_{\btheta} \log [p(\by \given \bX, \btheta) p(\btheta)]$
                \label{alg:bo:train}
                \Comment{\small{MAP estimate}}

            \State $\xnext \gets \argmax_{\bx} \alpha (\bx \given \by, \bX,\btheta)$
                \label{alg:bo:xnext}
                \Comment{\small{Maximise infill criterion}}
            \State $f' \gets f(\xnext)$
                \label{alg:bo:eval}
                \Comment{\small{Expensively evaluate $\xnext$}}

            \State $\bX \gets \bX \cup \{ \xnext \}$
                \label{alg:bo:augment}
                \Comment{\small{Augment training data}}
            
            \State $\by \gets \by \cup \{ f' \}$
        \EndFor
      \State \Return $\Data$
    \end{algorithmic}
  \end{algorithm}